%% file: sigconf.tex
\documentclass[sigconf]{acmart}

\usepackage{booktabs} 
\usepackage{multirow}
\usepackage{hyperref}
\usepackage{amsmath}

\copyrightyear{2018} 
\acmYear{2018} 
\setcopyright{acmcopyright}
\acmConference[MM '18]{2018 ACM Multimedia Conference}{October 22--26, 2018}{Seoul, Republic of Korea}
\acmBooktitle{2018 ACM Multimedia Conference (MM '18), October 22--26, 2018, Seoul, Republic of Korea}
\acmPrice{15.00}
\acmDOI{10.1145/3240508.3240552}
\acmISBN{978-1-4503-5665-7/18/10}

\begin{document}
\title{Learning Discriminative Features with Multiple Granularities \\ for Person Re-Identification}
\author{
	Guanshuo Wang$^{1*}$, 
	Yufeng Yuan$^{2*}$,
	Xiong Chen$^{2}$,
	Jiwei Li$^{2}$, 
	Xi Zhou$^{1,2}$
}
\affiliation{%
\institution{$^1$Cooperative Medianet Innovation Center, Shanghai Jiao Tong University}}
\affiliation{%
\institution{$^2$CloudWalk Technology}}
\email{guanshuo.wang@sjtu.edu.cn, {yuanyufeng,chenxiong,lijiwei,zhouxi}@cloudwalk.cn}
\thanks{*Equal contribution}
\renewcommand{\shortauthors}{G. Wang et al.}

\begin{abstract}
	The combination of global and partial features has been an essential solution to improve discriminative performances in person re-identification (Re-ID) tasks. Previous part-based methods mainly focus on locating regions with specific pre-defined semantics to learn local representations, which increases learning difficulty but not efficient or robust to scenarios with large variances. In this paper, we propose an end-to-end feature learning strategy integrating discriminative information with various granularities. We carefully design the Multiple Granularity Network (MGN), a multi-branch deep network architecture consisting of one branch for global feature representations and two branches for local feature representations. Instead of learning on semantic regions, we uniformly partition the images into several stripes, and vary the number of parts in different local branches to obtain local feature representations with multiple granularities. Comprehensive experiments implemented on the mainstream evaluation datasets including Market-1501, DukeMTMC-reid and CUHK03 indicate that our method robustly achieves state-of-the-art performances and outperforms any existing approaches by a large margin. For example, on Market-1501 dataset in single query mode, we obtain a top result of Rank-1/mAP=96.6\%/94.2\% with this method after re-ranking.
\end{abstract}

%
%

\ccsdesc[500]{Computing methodologies~Object identification}
\ccsdesc[500]{Computing methodologies~Learning to rank}
\ccsdesc[500]{Computing methodologies~Supervised learning by classification}
\ccsdesc[300]{Computing methodologies~Image representations}
\ccsdesc[100]{Computing methodologies~Neural networks}

\keywords{Person re-identification, Feature learning, Multi-branch deep network}

\maketitle

\input{body-conf}

\bibliographystyle{ACM-Reference-Format}
\bibliography{sample-bibliography}

\end{document}

%% file: body-conf.tex
\section{Introduction}
Person re-identification (Re-ID) is a challenging task to retrieve a given person among all the gallery pedestrian images captured across different security cameras. Due to the scene complexity of images from surveillance videos, the main challenges for person Re-ID come from large variations on persons such as pose, occlusion, clothes, background clutter, detection failure, etc. The prosperity of deep convolutional network has introduced more powerful representations with better discrimination and robustness for pedestrian images, which pushed the performance of Re-ID to a new level. Some recent deep Re-ID methods \cite{sarfraz2017pose,sun2017beyond,chang2018multilevel,shen2018deep,li2018harmoniou,shen2018end,chen2018group} have achieved breakthrough with high-level identification rates and mean average precision. 

\begin{figure}
	\includegraphics[width=3.35in]{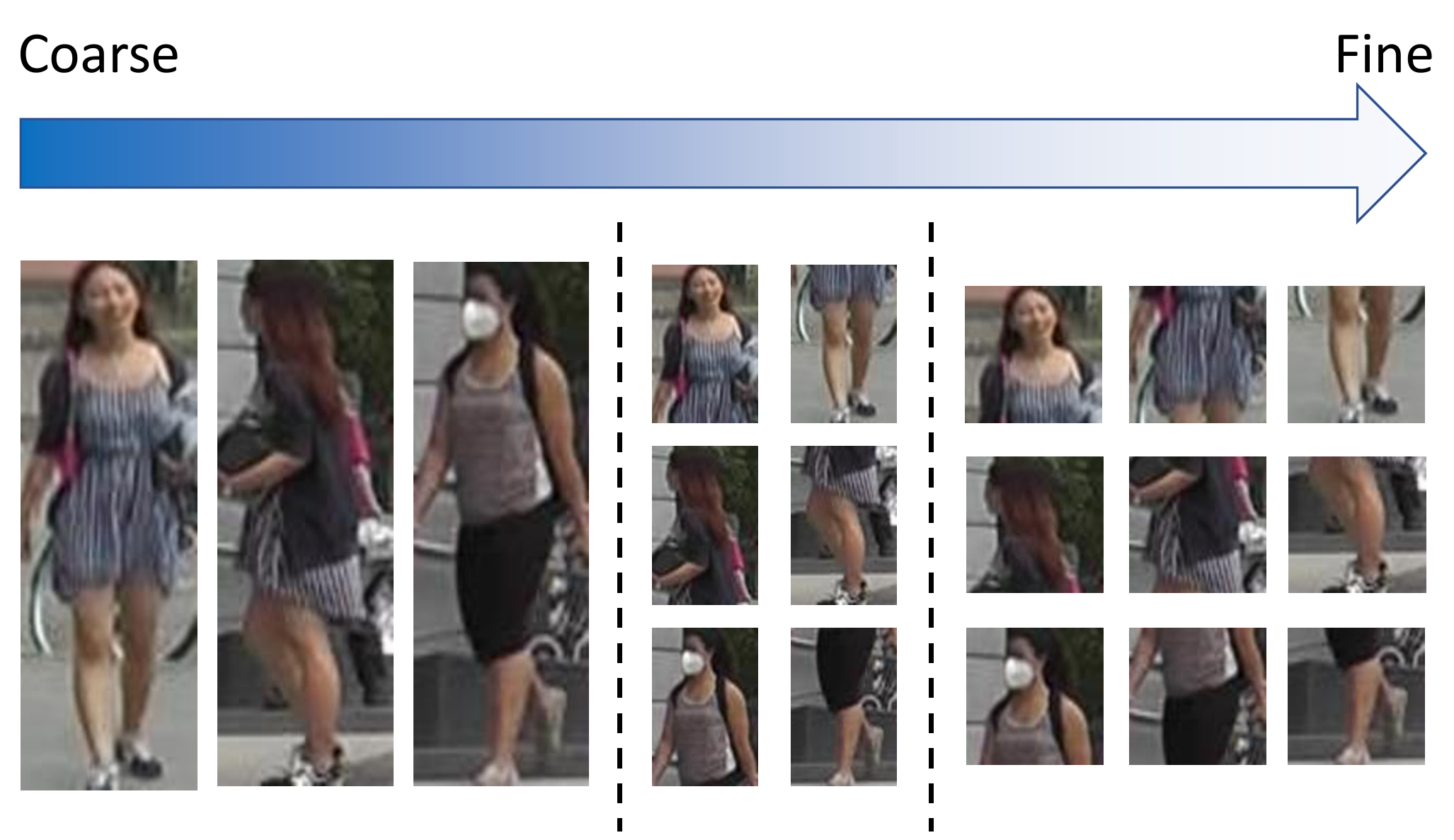}
	\caption{Body part partitions from coarse to fine granularities. We regard original pedestrian images with the whole body as the coarsest level of granularity in the left column. The middle and right column are respectively pedestrian partitions divided into 2 and 3 stripes from the original images. The more stripes images are divided into, the finer the granularity of partitions is.
	}
\end{figure}

The intuitive approach of pedestrian representations is to extract discriminative features from the whole body on images. The aim of global feature learning is to capture the most salient clues of appearance to represent identities of different pedestrians. However, high complexities for images captured in surveillance scenes usually restrict the accuracy for feature learning in large scale Re-ID scenarios. Due to the limited scale and weak diversity of person Re-ID training datasets, some non-salient or infrequent detailed information can be easily ignored and make no contribution for better discrimination during global feature learning procedure, which makes global features hard to adapt similar inter-class common properties or large intra-class differences. 

To relieve this dilemma, locating significant body parts from images to represent local information of identities has been confirmed to be an effective approach for better Re-ID accuracy in many previous works. Each located body part region only contains a small percentage of local information from the whole body, and at the same time distraction by other related or unrelated information outside the regions is actually filtered by locating operations, with which local features can be learned to concentrate more on identities and used as an important complement for global features. Part-based methods for person Re-ID can be divided into three main pathways according to their part locating methods: 1) Locating part regions with strong structural information such as empirical knowledge about human bodies \cite{cheng2016perso,li2017person,sun2017beyond,zhang2017alignedreid} or strong learning-based pose information \cite{su2017pose,zhao2017spindle}; 2) Locating part regions by region proposal methods \cite{yao2017deep,li2017learning}; 3) Enhancing features by middle-level attention on salient partitions \cite{zhao2017deeply,liu2017hydraplus,liu2017end,li2018harmoniou}. However, obvious limitations impede the effectiveness of these methods. First, pose or occlusion variations can affect the reliability of local representation. Second, these methods almost only focus on specific parts with fixed semantics, but cannot cover all the discriminative information. Last but not least, most of these methods are not end-to-end learning process, which increases the complexity and difficulty of feature learning.

In this paper, we propose a feature learning strategy combining global and local information in different granularities. As shown in Figure 1, various numbers of partition stripes introduce a diversity of content granularity. We define the original image containing only one whole partition with global information as the coarsest case, and as the number of partitions increase, features of local parts can concentrate more on finer discriminative information in each part stripe, filtering information on the other stripes. Since deep learning mechanism can capture approximate response preferences on the main body from the whole image, it is also possible to capture more fine-grained saliency for local features extracted from smaller part regions. Notice that these part regions are not necessary to be located partitions with specific semantics, but only a piece of equally-split stripe on the original images. From the observation, we find that the granularity of discriminative responses indeed becomes finer as the number of horizontal stripes increases. Based on this motivation, we design the Multiple Granularity Network (MGN), a multi-branch network architecture divided into one global and two local branches with delicated parameters from the 4th residual stage of the ResNet-50 \cite{he2016deep} backbone. In each local branch of MGN, we divide globally-pooled feature maps into different numbers of stripes as part regions to learn local feature representations independently, referring the methods in \cite{sun2017beyond}. 

Comparing to the previous part-based methods, our method only utilize equally-divided parts for local representation, but can achieve outstanding performance exceeding all previous methods. Besides, our method is completely a end-to-end learning process, which is easy for learning and implementation.  Extensive experiment results show that our method can achieve state-of-the-art performances on several mainstream Re-ID datasets, even with settings without any additional external data or re-ranking \cite{zhong2017rerank} operation.

\begin{figure}
	\includegraphics[width=2.8in]{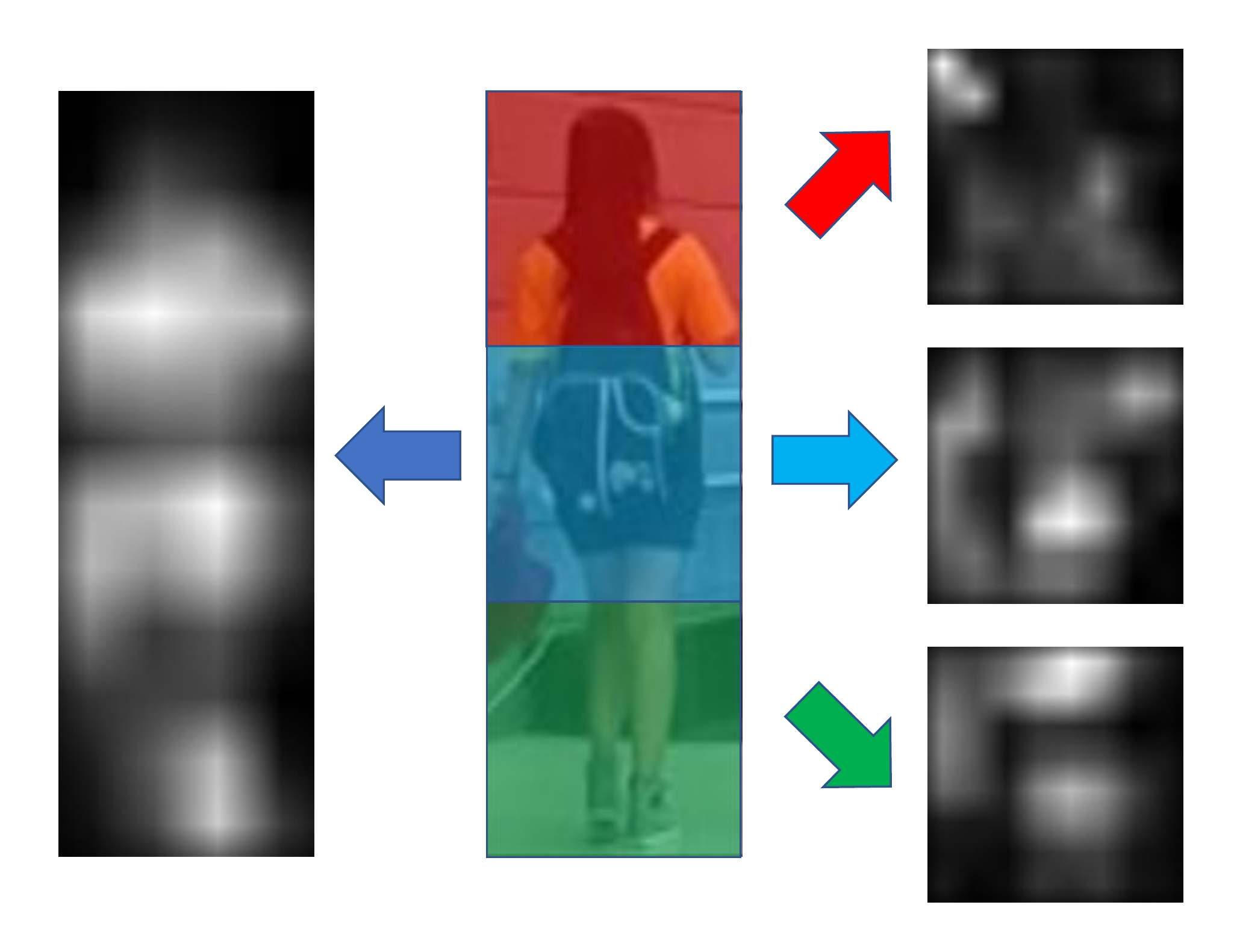}
	\caption{Feature response maps in different granularities extracted from the last output of different models. The response intensity is calculated by the L2-norm of feature vectors from all the spatial locations. Middle Column: a pedestrian image. Left Column: global response map by IDE embedding. Right Column: three local response maps corresponding to three split stripes of the origin image, extracted by part-based model. Best viewed in color.}
\end{figure}

\begin{figure*}
	\includegraphics[width=6in]{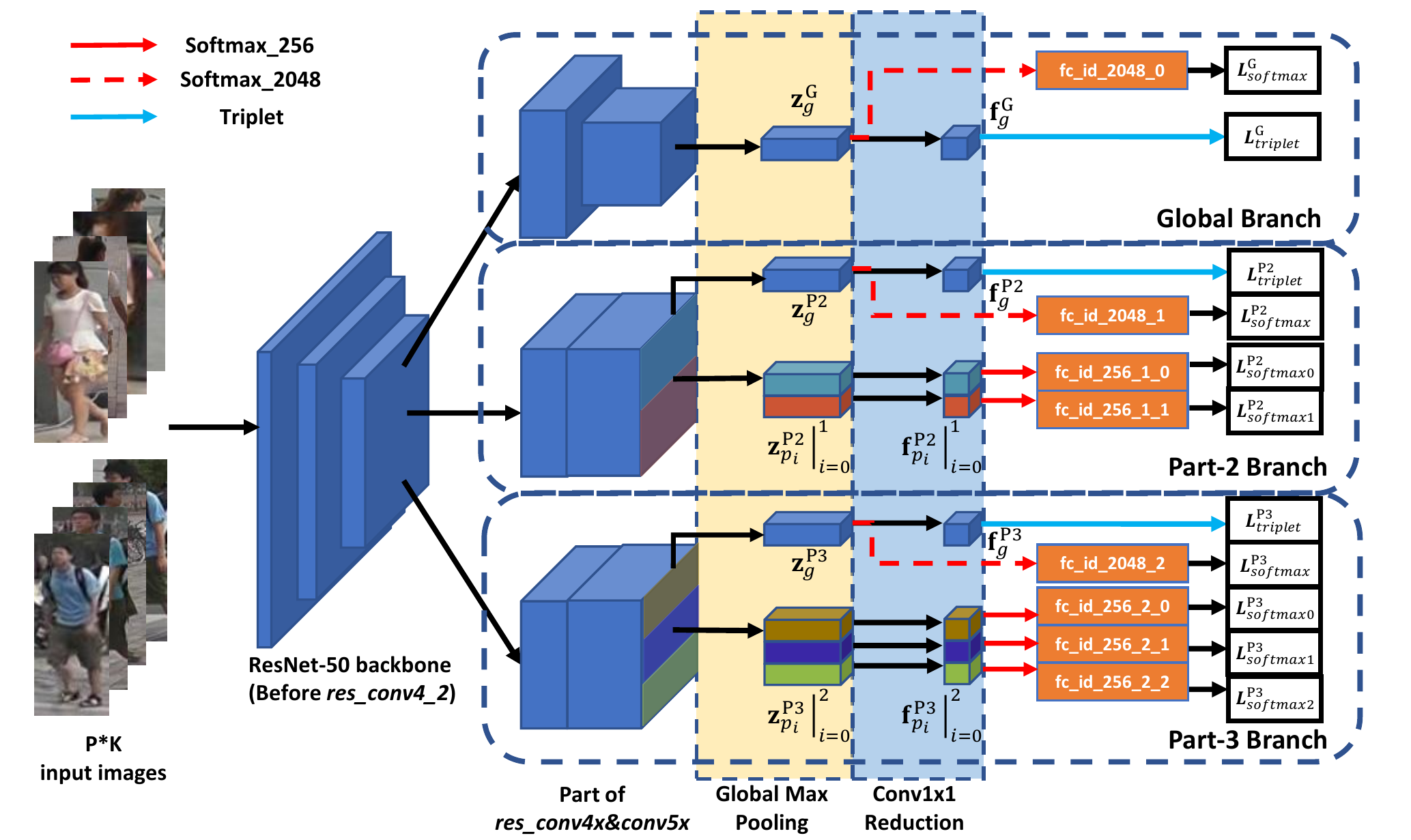}
	\caption{Multiple Granularity Network architecture. The ResNet-50 backbone is split into three branches after \textit{res\_conv4\_1} residual block: Global Branch, Part-2 Branch and Part-3 Branch. During testing, all the reduced features are concatenated together as the final feature representation of a pedestrian image. Notice that the $1 \times 1$ convolutions for dimension reduction and fully connected layers for identity prediction in each branch \textbf{DO NOT} share weights with each other. Each path from the feature to the specific loss function represents an independent supervisory signal. Best viewed in color.}
\end{figure*}

\section{Related Works}
With the prosperity of deep learning, feature learning by deep networks has become a common practice in person Re-ID tasks. \cite{li2014deepreid,yi2014deep} first introduce deep siamese network architecture into Re-ID and combine the body part feature learning, achieving higher performances comparing to the contemporary hand-crafted methods. \cite{zheng2016person} proposes ID-discriminative Embedding (IDE) with simple ResNet-50 backbones as a baseline of the performance level for modern deep Re-ID systems. A number of methods are proposed to improve the performance for deep person Re-ID. In \cite{ahmed2015improved, varior2016gated}, mid-level features of image pairs are computed to depict interrelation of local parts with carefully designed mechanism. \cite{xiao2016learnin} introduces Domain Guided Dropout to enhance the generalization ability across different domains of pedestrian scenarios. \cite{zhong2017rerank} brings re-ranking strategy into Re-ID tasks to modify the ranking results for accuracy improvement. 

Recently some deep Re-ID methods pushed the performances to a new level comparing to the former systems. \cite{zhang2017alignedreid} introduces a part-based alignment matching in training phase with shortest path programming and mutual learning to improve metric learning performance. \cite{bai2017deep, sun2017beyond} both equally slice the feature maps of input images into several stripes in vertical orientation. \cite{bai2017deep} merges slices of local features with LSTM network and combine with global features learned from classification metric learning. Instead \cite{sun2017beyond} directly concatenates the features from local parts as the final representation, and applies refined part pooling to modify the mapping validation of part features. However, according to the report in \cite{zhang2017alignedreid}, these systems just achieve similar performances as human, which we still need a highway to surpass. 

Among all the strategies for performance improvement, we argue that combining the local representations from parts of images is the most effective. As mentioned in Section 1, we summarize three main pathways for part-based learning: determining regions according to structural information about human body, locating body parts by region proposal methods and enhancing features by spatial attention. In \cite{cheng2016perso,li2017person,sun2017beyond}, images are all split into several stripes in horizontal orientation according to intrinsic human body structure knowledge, on which local feature representations are learned. \cite{zhao2017spindle,su2017pose} utilize structural information of body landmarks predicted by pose estimation methods to crop more accurate region areas with semantics. To locate semantic partitions without strongly learning-based predictors, region proposal methods such as \cite{girshick2015fast,jaderberg2015spatial} are employed in some part-based methods \cite{liu2017hydraplus,yao2017deep,zhao2017deeply,li2017learning,li2018harmoniou}. Attention information can be a powerful complement for discrimination, which are enhanced in \cite{liu2017end,liu2017hydraplus,li2018harmoniou}. In our proposed method, we only use simple horizontal stripes as part regions for local feature learning but achieve outstanding performances. 

Loss functions are used as supervisory signals in feature learning. In the training phase for deep Re-ID systems, the most common loss functions are classification losses and metric losses. Softmax loss is almost the only choice of classification loss function for its strong robustness to various kinds of multi-class classification tasks, which can be used individually \cite{ahmed2015improved,zheng2016person,xiao2016learnin,sun2017beyond,liu2017hydraplus,yao2017deep,li2017learning,li2018harmoniou} or combined with other losses \cite{li2014deepreid,cheng2016perso,zhang2017alignedreid,bai2017deep} in embedding learning procedures for Re-ID. For metric losses used in embedding learning for Re-ID, there are more variants with different ranking metrics. Contrastive loss \cite{hadsell2006dimensionality} is commonly used in siamese-liked networks \cite{varior2016gated}, which focuses on maximizing the distances between inter-class pairs and minimizing that between intra-class pairs. Triplet loss \cite{hoffer2015deep,schroff2015facenet} enforces a margin between the intra and inter distances to the same anchor sample with in a triplet. Based on triplet loss, many variants \cite{cheng2016perso,hermans2017defense,chen2017beyond,song2016deep} are proposed to solve learning or performance issues in metric learning. We employ a setting of joint learning with both softmax and triplet losses in our proposed method.


\section{Multiple Granularity Network}
Figure 2 shows feature response maps of a certain image extracted from the IDE baseline model \cite{zheng2016person} and a part-based model based on IDE. We can observe that even if no explicit attention mechanisms are imposed to enhance the preferences to some salient components, the deep network can still learn the preliminary distinction of response preferences on different body parts according to their inherent semantic meanings. However, to eliminate the distraction of unrelated patterns in pedestrian images with high complexity, higher responses are just concentrated on the main body of pedestrians instead of any concrete body parts with semantic patterns. As we narrow the area of represented regions and train as a classification task to learn local features, we can observe that responses on the local feature maps starts to cluster on some salient semantic patterns, which also varies with the sizes of represented regions. This observation reflects a relationship between the volume of image contents, \textit{i.e.} the granularity of regions, and capability of deep networks to focus on specific patterns for representations. We believe this phenomenon comes from the limitation of information in restricted regions. In general, comparing from a global image, it is intuitively hard to discriminate the identities for pedestrians from a local part region. Supervising signals of the classification task enforce the features to be correctly classified as the target identity, which also push the learning procedure trying to explore useful fine-grained details among limited information. 

Actually, local feature learning in previous part-based methods only introduces a basic granularity diversity of partitions to the total feature learning procedure with or without empirical prior knowledge. Assume that there are appropriate levels of granularities, details with most discriminative information might be almost concentrated by deep networks. Motivated by the observation and analysis above, we propose the Multiple Granularity Network (MGN) architecture to combine global and multi-granularity local feature learning for more powerful pedestrian representation. 

\begin{table} \footnotesize 
	\begin{tabular}{ccccc}
		\toprule
		Branch & Part No. & Map Size & Dims & Feature \\ 
		\midrule
		Global & 1 & $12\times4$ & 256 & $\mathbf{f}_g^{G}$\\
		Part-2 & 2 & $24\times8$ & 256*2+256 & $\{\mathbf{f}_{p_i}^{P2}|_{i=0}^1\}, \mathbf{f}_g^{P2}$\\
		Part-3 & 3 & $24\times8$ & 256*3+256 & $\{\mathbf{f}_{p_i}^{P3}|_{i=0}^2\}, \mathbf{f}_g^{P3}$\\
		\bottomrule
	\end{tabular}
	\caption{Comparison of the settings for three branches in MGN. Notice that the size of input images is set to $384\times128$. "Branch" refers to the name of branches. "Part No." refers to the number of partitions on feature maps. "Map Size" refers to the size of output feature maps from each branch. "Dims" refers to the dimensionality and number of features for the output representations. "Feature" means the symbols for the output feature representations.}
\end{table}

\subsection{Network Architecture}
The architecture of Multiple Granularity Network is shown in Figure 3. The backbone of our network is ResNet-50 which helps to achieve competitive performances in some Re-ID systems \cite{zhang2017alignedreid,bai2017deep,sun2017beyond}. The most obvious modification different from the original version is that we divide the subsequent part after \textit{res\_conv4\_1} block into three independent branches, sharing the similar architecture with the original ResNet-50.

Table 1 lists the settings of these branches. In the upper branch, we employ down-sampling with a stride-2 convolution layer in \textit{res\_conv5\_1} block, following a global max-pooling (GMP) \cite{almazan2018red} operation on the corresponding output feature map and a $1 \times 1$ convolution layer with batch normalization \cite{ioffe2015batch} and ReLU to reduce 2048-dim features $\mathbf{z}_g^G$ to 256-dim $\mathbf{f}_g^G$. This branch learns the global feature representations without any partition information, so we name this branch as the \textbf{Global Branch}.

The middle and lower branches both share the similar network architecture with Global Branch. The difference is that we employ no down-sampling operations in \textit{res\_conv5\_1} block to preserve proper areas of reception fields for local features, and output feature maps in each branch are uniformly split into several stripes in horizontal orientation, on which we independently perform the same following operations as Global Branch to learn local feature representations. We call these branches \textbf{Part-\textit{N} Branch}, where \textit{N} refers to the number of partitions on the unreduced feature maps, \textit{e.g.} the middle and lower branches in Figure 3 can be named as Part-2 and Part-3 Branch. 

During testing phases, to obtain the most powerful discrimination, all the features reduced to 256-dim are concatenated as the final feature, combining both the global and local information to perfect the comprehensiveness for learned features.

\subsection{Loss Functions}
To unleash the discrimination ability of the learned representations of this network architecture, we employ softmax loss for classfication, and triplet loss for metric learning as the loss functions in training phases, which are both widely used in various deep Re-ID methods. 

For basic discrimination learning, we regard the identification task as a multi-class classfication problem. For $i$-th learned features $\mathbf{f}_i$, softmax loss is formulated as:

\begin{equation}
L_{softmax}=-\sum_{i=1}^N{\log{\frac{e^{\mathbf{W}_{y_i}^T\mathbf{f_{i}}}}{\sum_{k=1}^{C}{e^{\mathbf{W}_{k}^T\mathbf{f_{i}}}}}}}
\end{equation}
where $\mathbf{W}_k$ corresponds to a weight vector for class $k$, with the size of mini-batch in training process \textit{N} and the number of classes in the training dataset \textit{C}. Different from the traditional softmax loss, the form we employ here abandons bias terms in linear multi-class classifiers according to \cite{wang2017normface}, which contributes to better discrimination performances. Among all the learned embeddings, we employ the softmax loss to the global features before $1 \times 1$ convolution reduction $\{\mathbf{z}_g^G, \mathbf{z}_g^{P2}, \mathbf{z}_g^{P3}\}$ and part features after reduction $\{\mathbf{f}_{p_i}^{P2}|_{i=1}^2, \mathbf{f}_{p_i}^{P3}|_{i=1}^3\}$.

All the global features after reduction $\{\mathbf{f}_g^G, \mathbf{f}_g^{P2}, \mathbf{f}_g^{P3}\}$ are trained with triplet loss to enhance ranking performances. We use the batch-hard triplet loss \cite{hermans2017defense}, an improved version based on the original semi-hard triplet loss. This loss function is formulated as follows:
\begin{equation}
\begin{aligned}
L_{triplet}=-\sum_{i=1}^P\sum_{a=1}^K [\alpha&+\max_{p=1...K}\|\textbf{f}_a^{(i)}-\textbf{f}_p^{(i)}\|_2\\&-\min_{\substack{n=1...K \\ j=1...P \\ j \neq i}}\|\textbf{f}_a^{(i)}-\textbf{f}_n^{(j)}\|_2]_+
\end{aligned}
\end{equation}
where $\textbf{f}_a^{(i)}, \textbf{f}_p^{(i)},\textbf{f}_n^{(i)}$ are the features extracted from anchor, positive and negative samples receptively, and $\alpha$ is the margin hyperparameter to control the differences of intra and inter distances. Here positive and negative samples refer to the pedestrians with same or different identity with the anchor. The candidate triplets are built by the furthest positive and closest negative sampled pairs, \textit{i.e.} the hardest positive and negative pairs in a mini-batch with $P$ selected identities and $K$ images from each identity. This improved version of triplet loss enhances the robustness in metric learning, and further improve the performances at the same time.  

In MGN achitecture, to avoid loss weight tuning troubles and difficulties in convergence, we novelly propose classfication-before-metric achitecture, which applies the softmax losses to reduced 256-dim local features in Part-2 and Part-3 Branches, and all the non-reduced global-pooled 2048-dim global features, but applies triplet losses to all the reduced features, different from existing methods using triplet losses. This setting is inspired from coarse-to-fine mechanism, regarding non-reduced features as coarse information to learn classification and reduced features as fine information with learned metric. The proposed setting achieves robust convergence comparing to that of imposing joint effects at the same level of reduced features. Besides, we employ no triplet loss on local features. Due to misalignment or other issues, the contents of local regions might vary dramatically, which makes the triplet loss tend to corrupt the model during training.

\subsection{Discussions}
In our proposed Multiple Granularity Network architecture, there are some issues worth our separate discussion. In this paragraph, we specifically discuss the issues as follows: 

\textbf{Multi-branch architecture} According to our initial motivation for MGN architecture, it seems to be reasonable that the global and local representations are both learned in one single branch. We can directly split the same final feature maps extracted by \textit{res\_conv5\_3} in different numbers of stripes, and apply corresponding supervisory signals as our proposed methods. However, we find this setting is not efficient for further performance improving. Borrowing the ideas in \cite{sun2014deeply}, the reason might be that the branches sharing the similar network architecture(mainly the fourth residual stage of ResNet-50) just response to different levels of detailed information on images. Learning features in multiple granularities with one mixed single branch might dilute the importance of detailed information. Besides, we try to split the backbone network after shallower or deeper layers, which also achieve no better performances. 

\textbf{Diversity of granularity} Three branches in our network architecture actually learn representing information with different perferences. Global Branch with larger reception field and global max-pooling captures integral but coarse features from the pedestrian images, and features learned by Part-2 and Part-3 Branches without strided convolution and split parts of stripes tend to be local but fine. The branch with more partitions will learn finer representation for pedestrian images. Branches learning different preferences can cooperatively supplement low-level discriminating information to the common backbone parts, which is the reason for performance boosting in any single branch. 

\section{Experiment}
\subsection{Implementation}
To capture more detailed information from pedestrian images, we refer to \cite{sun2017beyond} and resize input images to $384 \times 128$.  We use the weights of ResNet-50 pretrained on ImageNet \cite{deng2009imagenet} to initialize the backbone and branches of MGN. Notice that different branches in the network are all initialized with the same pretrained weights of the corresponding layers after the \textit{res\_conv4\_1} block. During training phases, we only deploy random horizontal flipping to images in the training dataset for data augmentation. Each mini-batch is sampled with randomly selected \textit{P} identities and randomly sampled \textit{K} images for each identity from the training set to cooperate the requirement of triplet loss. Here we recommend to set $P=16$ and $K=4$ to train our proposed model. For the margin parameter for triplet loss, we set to 1.2 in all our experiments. We choose SGD as the optimizer with momentum 0.9. The weight decay factor for L2 regularization is set to 0.0005. As for the learning rate strategy, we set the initial learning rate to 0.01, and decay the learning rate to 1e-3 and 1e-4 after training for 40 and 60 epochs. The total training process lasts for 80 epochs. During evaluation, we both extract the features corresponding to original images and the horizontally flipped versions, then use the average of these as the final features. Our model is implemented on PyTorch framework. To conduct a complete training procedure on Market-1501 dataset, it takes about 2 hours with data-parallel acceleration by two NVIDIA TITAN Xp GPUs. All our experiments on different datasets follow the settings above.

\subsection{Datasets and Protocols}
The experiments to evaluate our proposed method are conducted on three mainstream Re-ID datasets: Market-1501 \cite{zheng2015scalable}, DukeMTMC-reID \cite{zheng2017unlabeled} and CUHK03 \cite{li2014deepreid}. It is necessary to introduce these datasets and their evaluation protocols before we show our results.

\textbf{Market-1501} This dataset includes images of 1,501 persons captured from 6 different cameras. The pedestrians are cropped with bounding-boxes predicted by DPM detector \cite{felzenszwalb2008discriminatively}. The whole dataset is divided into training set with 12,936 images of 751 persons and testing set with 3,368 query images and 19,732 gallery images of 750 persons. There are single-query and multiple-query modes in evaluation, the difference of which is the number of images from the same identity. In multiple-query mode, all features extracted from the images of a person captured by the same camera are merged by avg- or max-pooling, which contains more complete information than single query mode with only 1 query image. 

\textbf{DukeMTMC-reID} This dataset is a subset of the DukeMTMC \cite{ristani2016MTMC} used for person re-identification by images. It consists of 36,411 images of 1,812 persons from 8 high-resolution cameras. 16,522 images of 702 persons are randomly selected from the dataset as the training set, and the remaining 702 persons are divided into the testing set where contains 2,228 query images and 17,661 gallery images. It might be the most challenging datasets for person Re-ID at present, with common situations in high similarity across persons and large variations within the same identity.

\textbf{CUHK03} This dataset consists of 14,097 images of 1,467 persons from 6 cameras. Two types of annotations are provided in this dataset: manually labeled pedestrian bounding boxes and DPM-detected bounding boxes. Originally the whole dataset is divided into 20 random splits for cross-validation, which is designed for hand-crafted methods and very time-consuming to conduct experiments for deep-learning-based methods. 

\textbf{Protocols} In our experiments, to evaluate the performances of Re-ID methods, we report the cumulative matching characteristics (CMC) at rank-1, rank-5 and rank-10, and mean average precision (mAP) on all the candidate datasets. On Market-1501 dataset, we conduct experiments both in single-query and multiple-query mode. On CUHK03 dataset, to simplify the evaluation procedure and meanwhile enhance the accuracy of the performance reflected by the results, we adopt the protocol used in \cite{zhong2017rerank}.

\begin{table} \small
	\label{tab:comparison_market1501}
	\centering
	\begin{tabular}{c|cc|cc}
		\toprule
		\multirow{2}{*}{Methods} & \multicolumn{2}{c|}{Single Query} & \multicolumn{2}{c}{Multiple Query}\\
		& Rank-1 & mAP & Rank-1 & mAP \\ 
		\midrule
		TriNet\cite{hermans2017defense} & 84.9 & 69.1 & 90.5 & 76.4 \\
		JLML\cite{li2017person} & 85.1 & 65.5 & 89.7 & 74.5 \\
		AACN\cite{xu2018attention} & 85.9 & 66.9 & 89.8 & 75.1 \\
		AOS\cite{huang2018adversarially} & 86.5 & 70.4 & 91.3 & 78.3 \\
		DPFL\cite{chen2017person} & 88.6 & 72.6 & 92.2 & 80.4 \\
		MLFN\cite{chang2018multilevel} & 90.0 & 74.3 & 92.3 & 82.4 \\
		KPM+RSA+HG\cite{shen2018end} & 90.1 & 75.3 & - & - \\
		PSE+ECN\cite{sarfraz2017pose} & 90.4 & 80.5 & - & - \\
		HA-CNN\cite{li2018harmoniou} & 91.2 & 75.7 & 93.8 & 82.8 \\
		DuATM\cite{si2018dual} & 91.4 & 76.6 & - & - \\
		GSRW\cite{shen2018deep} & 92.7 & 82.5 & - & - \\
		DNN\_CRF\cite{chen2018group} & 93.5 & 81.6 & - & - \\
		PCB+RPP\cite{sun2017beyond} & 93.8 & 81.6 & - & - \\
		MGN(Ours) & \textbf{95.7} & \textbf{86.9} & \textbf{96.9} & \textbf{90.7} \\
		\midrule
		TriNet(RK)\cite{hermans2017defense} & 86.7  & 81.1 & 91.8 & 87.2 \\
		AOS(RK)\cite{huang2018adversarially} & 88.7 & 83.3 & 92.5 & 88.6 \\
		AACN(RK)\cite{xu2018attention} & 88.7 & 83.0 & 92.2 & 87.3 \\
		PSE+ECN(RK)\cite{sarfraz2017pose} & 90.3 & 84.0 & - & - \\
		MGN(Ours, RK) & \textbf{96.6} & \textbf{94.2} & \textbf{97.1} & \textbf{95.9}\\
		\bottomrule
	\end{tabular}
	\caption{Comparison of results on Market-1501 with Single Query setting (SQ) and Multiple Query setting (MQ). "RK" refers to implementing re-ranking operation.}
\end{table}

\begin{table}
	\centering
	\label{tab:comparison_duke}
	\begin{tabular}{c|cc}
		\toprule
		Methods & Rank-1 & mAP \\
		\midrule
		SVDNet\cite{sun2017svdnet} & 76.7 & 56.8 \\
		AOS\cite{huang2018adversarially} & 79.2 & 62.1 \\
		HA-CNN\cite{li2018harmoniou} & 80.5 & 63.8 \\
		GSRW\cite{shen2018deep} & 80.7 & 66.4 \\
		DuATM\cite{si2018dual} & 81.8 & 64.6 \\
		PCB+RPP\cite{sun2017beyond} & 83.3 & 69.2 \\
		PSE+ECN\cite{sarfraz2017pose} & 84.5 & 75.7 \\
		DNN\_CRF\cite{chen2018group} & 84.9 & 69.5 \\
		GP-reid\cite{almazan2018red} & 85.2 & 72.8 \\ 
		\midrule
		MGN(Ours) & \textbf{88.7} & \textbf{78.4} \\
		\bottomrule
	\end{tabular}
	\caption{Comparison of results on DukeMTMC-reID.}
\end{table}

\begin{table}
	\label{tab:comparison_cuhk03}
	\centering
	\begin{tabular}{c|cc|cc}
		\toprule		\multirow{2}{*}{Methods} & \multicolumn{2}{c|}{Labeled} & \multicolumn{2}{c}{Detected}\\
		& Rank-1 & mAP & Rank-1 & mAP \\
		\midrule
		BOW+XQDA\cite{zheng2015scalable} & 7.9 & 7.3 & 6.4 & 6.4 \\
		LOMO+XQDA\cite{liao2015person} & 14.8 & 13.6 & 12.8 & 11.5 \\
		\midrule
		IDE\cite{zheng2016person} & 22.2 & 21.0 & 21.3 & 19.7 \\
		PAN\cite{zheng2017pedestrian} & 36.9 & 35.0 & 36.3 & 34.0 \\
		SVDNet\cite{sun2017svdnet} & 40.9 & 37.8 & 41.5 & 37.3 \\
		HA-CNN\cite{li2018harmoniou} & 44.4 & 41.0 & 41.7 & 38.6 \\
		MLFN\cite{chang2018multilevel} & 54.7 & 49.2 & 52.8 & 47.8 \\
		PCB+RPP\cite{sun2017beyond} & - & - & 63.7 & 57.5 \\
		\midrule
		MGN(Ours) & \textbf{68.0} & \textbf{67.4} & \textbf{66.8} & \textbf{66.0} \\ 
		\bottomrule
	\end{tabular}
	\caption{Comparison of results on CUHK03 with evaluation protocols in \cite{zhong2017rerank}.}
\end{table}

\subsection{Comparison with State-of-the-Art Methods}
We compare our proposed method with current state-of-the-art methods on all the candidate datasets to show our considerable performance advantage over all the existing competitors. Results in detail are given as follow:

\textbf{Market-1501} The results on Market-1501 dataset is shown in Table 2. For the special effects of re-ranking method for improvement on mAP and rank-1 accuracy, we divide the results into two groups according to whether re-ranking is implemented or not. In single query mode, PCB+RPP \cite{sun2017beyond} achieved the best published result without re-ranking , but our MGN achieves Rank-1/mAP=95.7\%/86.9\%, exceeding the former method by 1.9\% in Rank-1 accuracy and 5.3\% in mAP. After implementing re-ranking, the result can be improved to Rank-1/mAP=96.6\%/94.2\%, which surpasses all existing methods by a large margin. 

Figure 4 shows top-10 ranking results for some given query pedestrian images. The first two results shows the great robustness: regardless of the pose or gait of these captured pedestrian, MGN features can robustly represent discriminative information of their identities. The third query image is captured in a low-resolution condition, losing an amount of important information. However, from some detailed clues such as the strap of the bag and his black suits, most of the ranking results are accurate and with high quality. The last pedestrian shows his back carrying a black backpack, but we can obtain his captured images in front view in rank-3, 6 and 9. We attribute this surprising result to the effects of local features, which establish relationships when some salient parts are lost. 

\textbf{DukeMTMC-reID} According to Table 3, our MGN architecture also performs excellently on the challenging DukeMTMC-reID dataset. GP-reid \cite{almazan2018red} is a good practice of many useful strategies combined in person Re-ID tasks and achieved the best published result. MGN achieves state-of-the-art result of Rank-1/mAP=88.7\%/78.4\%, outperforming GP-reid by +3.5\% in Rank-1 and +5.6\% in mAP. Standing on this level of performance on the most challenging datasets currently, we believe there are still some issues to be conquered for further perfect deep Re-ID systems. 

\textbf{CUHK03} As shown in Table 4, our MGN achieves Rank-1/mAP= 68.0\%/67.4\% on CUHK03 labeled setting and 66.8\%/66.0\% on CUHK03 detected setting, which outperform all the published results by a large margin. Here we can observe an obvious gap between results of labeled and detected conditions. We argue that it reflects an important affect of detection failure on person Re-ID performance, which emphasizes the importance of high-performance pedestrian detectors.

\begin{table} \small
	\label{tab:our_method}
	\centering
	\begin{tabular}{c|cccc}
		\toprule
		Model & Rank-1 & Rank-5 & Rank-10 & mAP \\ 
		\midrule
		ResNet-50 & 87.5 & 94.9 & 96.7 & 71.4 \\
		ResNet-101 & 90.4 & 95.7 & 97.2 & 78.0 \\ 
		ResNet-50+TP & 88.7 & 96.0 & 97.2 & 75.0 \\
		\midrule
		Global (Branch) & 89.8 & 95.8 & 97.5 & 78.5 \\
		Part-2 (Single) & 92.6 & 97.1 & 98.0 & 80.2 \\
		Part-2 (Branch) & 94.4 & 97.9 & 98.8 & 83.9 \\
		Part-3 (Single) & 93.1 & 97.6 & 98.7 & 82.1 \\
		Part-3 (Branch) & 94.4 & 98.2 & 98.8 & 84.1 \\
		G+P2+P3 (Single) & 94.4 & 97.6 & 98.5 & 85.2 \\
		\midrule
		MGN w/o Part-3 & 94.4 & 97.9 & 98.7  & 85.7  \\
		MGN w/ Part-4 & 95.1 & 98.3 & 98.9    & 86.1  \\
		\midrule
		MGN (Part2+4) & 94.8 & 98.2 & 98.9 & 85.6 \\
		MGN (Part3+4) & 95.0 & 98.1 & 98.8 & 86.1 \\
		\midrule
		MGN w/o TP & 95.3 & 97.9 & 98.7 & 86.2 \\
		MGN & \textbf{95.7} & \textbf{98.3} & \textbf{99.0} & \textbf{86.9}  \\
		\bottomrule
	\end{tabular}
	\caption{Results with different settings on Market-1501 datasets. "TP" refers to triplet loss. "Branch" refers to a sub-branch of MGN. "Single" refers to a single network with the same setting as the branch with the corresponding name in MGN. The model "ResNet-50+TP" can be regarded as "Global (Single)". "G+P2+P3" refers to an ensemble setting by Global (Single), Part-2 (Single) and Part-3 (Branch).}
\end{table}

\begin{figure*}
	\includegraphics[width=6in]{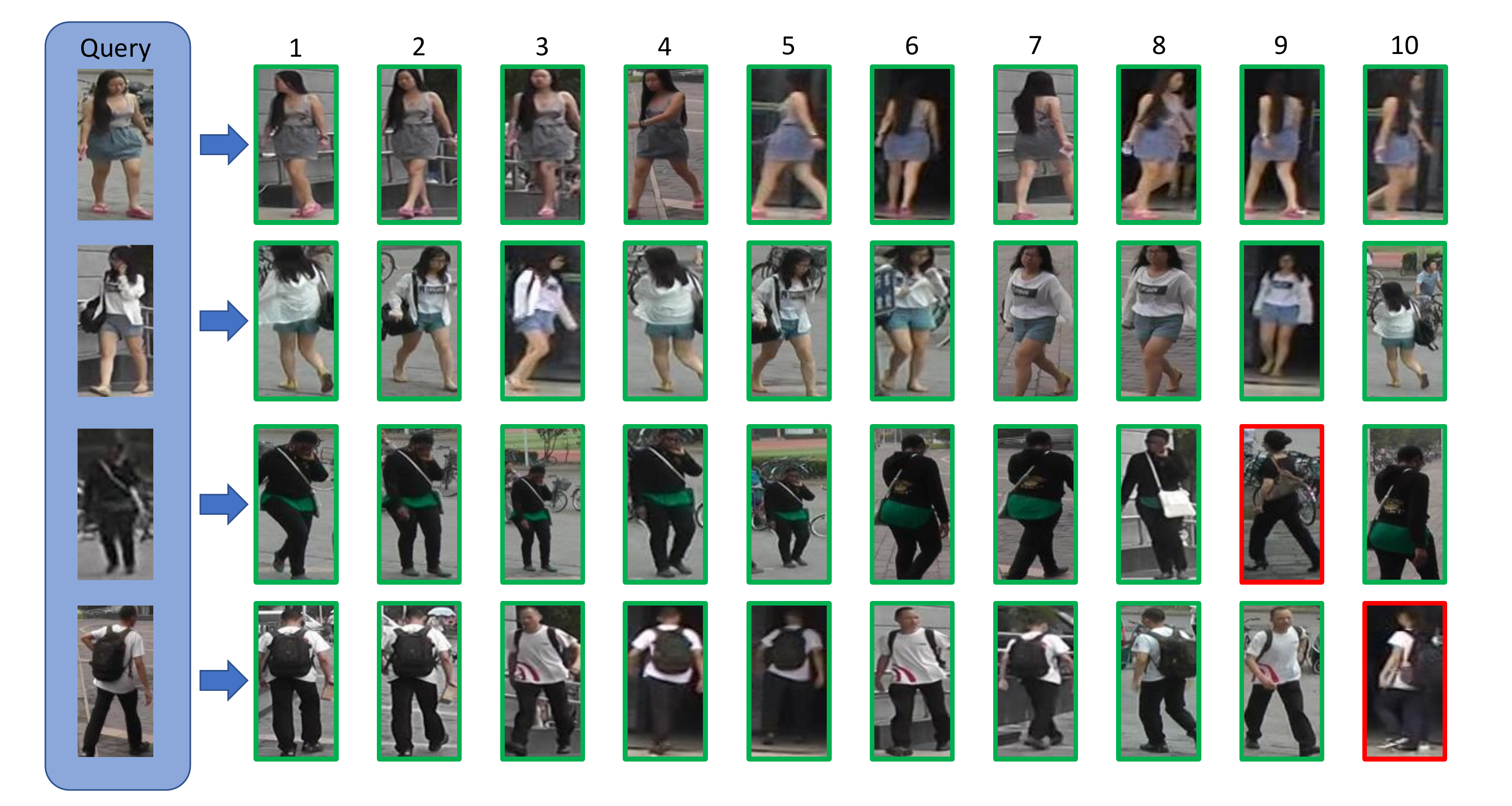}
	\caption{Top-10 ranking list for some query images on Market-1501 datasets by MGN. The retrieved images are all from in the gallery set, but not from the same camera shot. The images with green borders belong to the same identity as the given query, and that with red borders do not. }
\end{figure*}

\subsection{Effectiveness of Components}
To verify effectiveness of each component in MGN, we conduct several ablation experiments with different component settings on Market-1501 dataset in single query mode. Notice that other unrelated settings in each comparative experiment are the same as MGN implementation in Section 4.1, and we have carefully tuned all the candidate models and report the best performance with our settings. Table 5 shows the comparison results in different settings related to components of MGN. We separately analyze each component as follows: 

\textbf{MGN vs ResNet-50} Comparing the results by the baseline ResNet-50 model with our MGN model without triplet loss, we can observe MGN makes a significant performance improvement from Rank-1/mAP=87.5\%/71.4\% to 95.3\%/86.2\% (+7.8\%/14.8\%). We also implement the same experiment with ResNet-101 model, which has a similar scale of weights with MGN. The deeper ResNet-101 network indeed brings a considerable performance boost (+2.9\%/7.4\%), but there is still a large gap with our MGN model, which shows that extra weights from additional branches are not the main contributors of the improvement, but the carefully-designed network architecture. Results above prove that our proposed MGN has incredible capability of feature representations for person Re-ID.

\begin{figure}
	\includegraphics[width=2.3in]{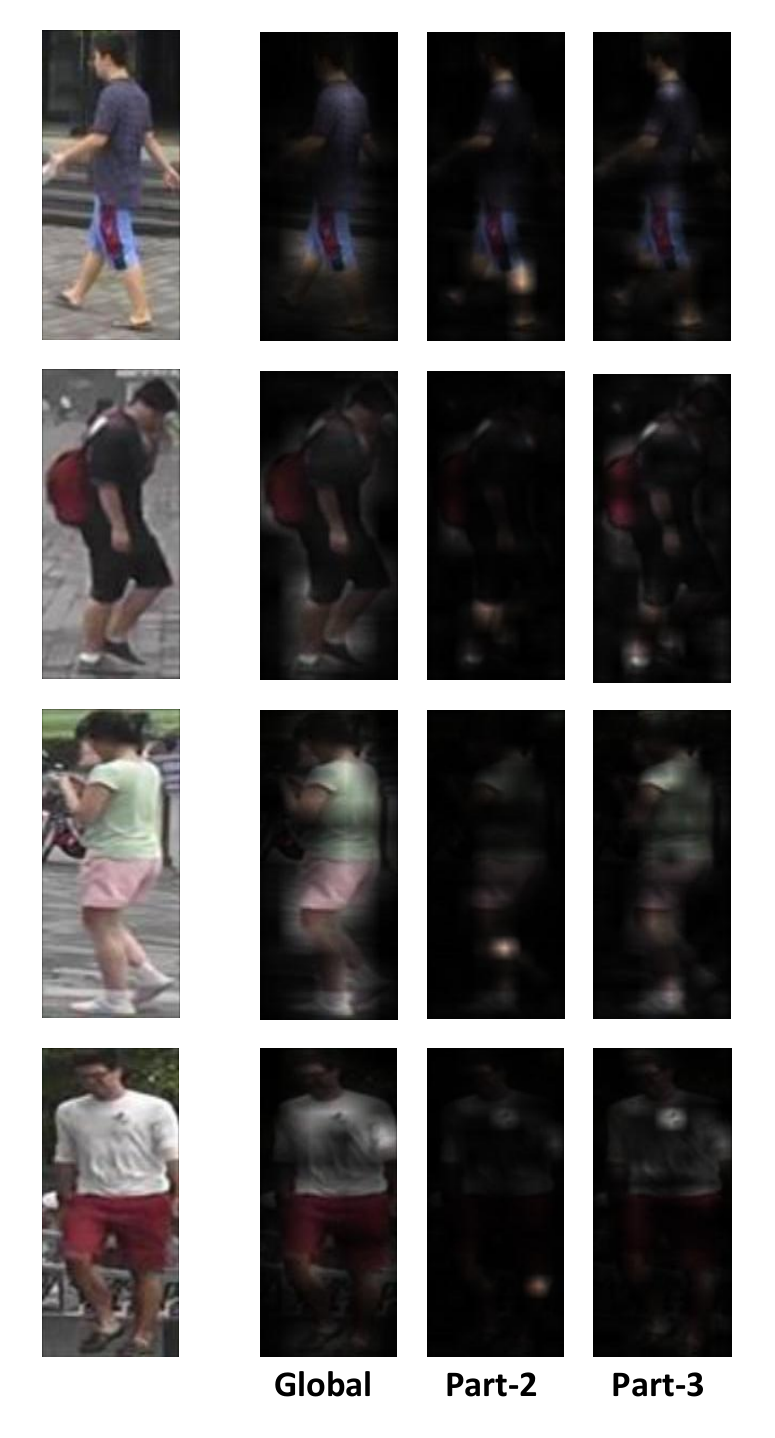}
	\caption{Feature response maps extracted from output layers of every branches. First column: the original pedestrian images. Second column: response maps from Global Branch. Third column: response maps from Part-2 Branch. Fourth column: response maps from Part-3 Branch. The brighter the area is, the more concentrated it is. Best viewed in color. }
\end{figure}

\textbf{Multiple branch vs Multiple networks} The multi-branch setting in one single network is very similar to ensemble of multiple independent networks, but we believe the cooperation of multiple branches can achieve a better performance than ensemble learning. We train three independent ResNet-50 networks separately, each of which respectively replicates the corresponding configuration of three branches, \textit{i.e.} Global, Part-2 and Part-3 Branch in MGN . In our experiments, we explore the effects of multi-branch architecture in two aspects. On the one hand, from a global view, we compare the performance of MGN with the ensemble of three single networks. The ensemble strategy indeed achieves better performance than any single participating network, but MGN still outperforms about $1\%\sim2\%$ on both Rank-1 and mAP. It shows that the cooperation of branches learns more discriminative feature representations than independent networks. On the other hand, from a local view, we respectively compare the performances of features learned by sub-branches of MGN with single networks in corresponding setting of branch. As our expect, the features from sub-branches also perform better than that from single networks. We argue that the mutual effects between sub-branches complement the blind spots in their individual learning procedure. 

\textbf{Multi-branch architecture settings} The multi-branch deep network architectures are very common for person Re-ID tasks \cite{cheng2016perso}. Our proposed MGN exceeds all the previous architectures by the power of learning representations with multiple granularities. Based on our proposed architecture, we can intuitively infer many variants architectures. On the one hand, based on the MGN model, we can add or reduce the number of local branches. Comparing the model removed the Part-3 branch with that added a Part-4 branch, we find that the removal of Part-3 Branch (MGN w/o Part-3) brings obvious performance degradation by Rank-1/mAP=$-$0.9\%/0.7\%, and the addition of Part-4 Branch (MGN w/ Part-4) introduce no obvious performance boost. This confirms the necessity and efficiency of our proposed 3-branch setting. 
On the other hand, for the Part-\textit{N} local branches, the number of partitions is a hyperparameter effecting the granularity of learning representations. We divide the feature maps into 2, 3, 4 in each local branches alternatively, and still find that the proposed Part-2/Part-3 setting is optimal. The Part-2/Part-4 setting skips the granularity level of Part-3, and brings performance degradation by Rank-1/mAP=$-$0.9\%/1.3\%, which shows the importance of Part-3 granularity. Comparing the results of Part-2/Part-4 setting with Part-3/Part-4, the former setting causes greater performance loss than the later. The reason is that the uniform division by 2 and 4 introduces no overlap areas between stripes, but the Part-3/Part-4 setting does. We argue this overlap can introduce the correlation between different partitions, from which more discriminative information can be learned.

\textbf{Triplet Loss} A number of previous works \cite{li2014deepreid,cheng2016perso,zhang2017alignedreid,bai2017deep} have shown the effectiveness of joint training with softmax loss for classification and triplet loss for metric learning in person Re-ID tasks. In our experiments, we reproduce the boosting effect with both ResNet-50 and MGN models. With the help of triplet loss on all the candidate datasets, we can observe +1.2\%/3.6\% Rank-1/mAP improvement to the baseline model, and +0.4\%/0.7\% Rank-1/mAP improvement to MGN model. We can observe two interesting effects from the improvement figures: 1) The improvement on mAP is more obvious than that on rank-1 accuracy, which proves the ranking effects of metric learning losses. 2) Triplet loss brings larger improvement to the baseline model than that to MGN. Comparing to softmax loss, triplet loss helps to capture more detailed information to meet the margin condition \cite{schroff2015facenet}. Our proposed network architecture is initially designed to enhance the local representation, which dilutes the original effects of triplet loss. Notice that in the MGN without triplet loss setting (MGN w/o TP), we employ no softmax loss on 2048-dim features, and alternatively on the reduced 256-dim features.

\textbf{Feature response with multiple granularity} We insist on our proposed MGN model learns the global and local feature representations with multiple levels of granularities. Figure 5 shows some feature response maps for some input pedestrian images, extracted from all the top of branches in MGN. All the response maps filter most of the complex background, which contain no useful information about identities for pedestrians. The responses from Global Branch is mainly focused on main body parts, and the information on limbs, waists or feet is commonly ignored. 
On local branches, the global responses on main body are lost, but concentrate more on some particular parts. For example, we can observe preferences on body parts such as shoulders and joints in Part-2 cases. As for the Part-3 cases, responses are more scattered on body parts, but some pivotal semantic information is preferred. Notice that on the Part-3 response map of the last pedestrian images, we can observe a bright circle region in front of the chest of this pedestrian. It is a mark of his T-shirt, which can be regarded as a very discriminative characteristic for his identity. 


\section{Conclusion}
In this paper, we propose the Multiple Granularity Network (MGN), a novel multi-branch deep network for learning discriminative representations in person re-identification tasks. Each branch in MGN learns global or local representation with certain granularity of body partition. Our method directly learns local features on horizontally-split feature stripes, which is completely end-to-end and introduces no part locating operations such as region proposal or pose estimation. Extensive experiments have indicated that our method not only achieves state-of-the-art results on several mainstream person Re-ID datasets, but also pushes the performance to an exceptional level comparing to existing methods.